\documentclass[10pt,twocolumn,letterpaper]{article}

\usepackage[pagenumbers]{wacv} 

\usepackage{times}
\usepackage{color}
\usepackage{colortbl}
\usepackage{epsfig}
\usepackage{graphicx}
\usepackage{amsmath}
\usepackage{eucal}
\usepackage{amssymb}
\usepackage{graphicx}
\usepackage{amsmath}
\usepackage{amssymb}
\usepackage{booktabs}
\usepackage{multirow,booktabs,makecell}

\usepackage[pagebackref,breaklinks,colorlinks]{hyperref}

\usepackage[capitalize]{cleveref}
\crefname{section}{Sec.}{Secs.}
\Crefname{section}{Section}{Sections}
\Crefname{table}{Table}{Tables}
\crefname{table}{Tab.}{Tabs.}

\def\wacvPaperID{1638} 
\def\confName{WACV}
\def\confYear{2024}

\definecolor{hzw_comment}{rgb}{.1,.1,.9} 

\definecolor{MyRed}{rgb}{0.8,0.2,0}
\definecolor{MyBlue}{rgb}{0,0,1.0}

\newcommand{\backwardwarp}{\overleftarrow{\mathcal{W}}}
\definecolor{xj}{rgb}{1,0.5,0}
\def\ie{\emph{i.e.}}
\def\etal{{\em et al.}}
\newcommand{\sota}{state-of-the-art}
\usepackage{cleveref}
\crefformat{section}{\S#2#1#3}
\crefformat{subsection}{\S#2#1#3}
\crefformat{subsubsection}{\S#2#1#3}
\crefformat{figure}{Figure~#2#1#3}
\crefformat{table}{Table~#2#1#3}

\newcommand{\citecomment}[1]{[\textcolor{green}{$\ast$}]{}}

\graphicspath{{appendix/}}

\begin{document}

\title{Scale-Adaptive Feature Aggregation for Efficient Space-Time Video Super-Resolution}
\newcommand*\samethanks[1][\value{footnote}]{\footnotemark[#1]}
\author{
		Zhewei Huang$^{1}$
        ~~~~
		Ailin Huang$^{1}$
        ~~~~        
        Xiaotao Hu$^{1,2}$
		~~~~
  Chen Hu$^{1}$
		~~~~
            Jun Xu$^{2,3,}$\thanks{Corresponding authors.}
        ~~~~
            Shuchang Zhou$^{1,}$\samethanks{}\\
        $^{1}$Megvii Technology~~~~
        $^{2}$Nankai University\\
        $^{3}$Guangdong Provincial Key Laboratory of Big Data Computing, CUHK\\
        \tt\small hzwer@pku.edu.cn, csjunxu@nankai.edu.cn, shuchang.zhou@gmail.com\\  
        \url{https://github.com/megvii-research/WACV2024-SAFA}
	}
 
\maketitle
\begin{abstract}
The Space-Time Video Super-Resolution (STVSR) task aims to enhance the visual quality of videos, by simultaneously performing video frame interpolation (VFI) and video super-resolution (VSR). However, facing the challenge of the additional temporal dimension and scale inconsistency, most existing STVSR methods are complex and inflexible in dynamically modeling different motion amplitudes. In this work, we find that choosing an appropriate processing scale achieves remarkable benefits in flow-based feature propagation. We propose a novel Scale-Adaptive Feature Aggregation (SAFA) network that adaptively selects sub-networks with different processing scales for individual samples. Experiments on four public STVSR benchmarks demonstrate that SAFA achieves state-of-the-art performance. Our SAFA network outperforms recent state-of-the-art methods such as TMNet~\cite{xu2021tmnet} and VideoINR~\cite{chen2022videoinr} by an average improvement of over 0.5dB on PSNR, while requiring less than half the number of parameters and only 1/3 computational costs. 
\end{abstract}

\section{Introduction}

Constrained by the filming, processing, and distribution pipelines, the majority of videos are commonly stored and displayed at limited resolution and frame rates. Within the sphere of industry, space-time video super-resolution~(STVSR) is practical for synthesizing smooth high-definition videos. New applications and software have continuously created demands for improving the performance and efficiency of STVSR models. From a scholarly perspective, the STVSR methods inspire insights into motion modeling for many video processing tasks~\cite{sevilla2019integration,wu2022optimizing}. 
Formally, given two low-resolution (LR) frames $\{I_0^{LR}, I_1^{LR}\}$ and a target time-step $t\in[0,1]$, our goal is to synthesize the high-resolution (HR) frame $I_t^{HR}$ at moment $t$. The STVSR task contains two correlated subtasks of video frame interpolation~(VFI) and video super-resolution~(VSR). Both are among the most studied problems in Computer Vision. A traditional approach to efficiently aggregate information across frames involves estimating the dense displacement field. This is represented as $\textbf{f}_{0\rightarrow 1}$ for a pair of frames $\{I_0, I_1\}$, known as the optical flow. This technique is widely used in VSR. Besides, most VFI methods~\cite{liu2017video,jiang2018super,bao2019depth,xu2019quadratic,huang2022rife} focus on approximating the backward flow fields $\{\textbf{f}_{t\rightarrow 0}, \textbf{f}_{t\rightarrow1}\}$ starting from the intermediate frame $I_t$. It is different from the former~(because $I_t$ is to be predicted) but very related. Successively performing VFI~\cite{bao2019depth,kong2022ifrnet} and VSR~\cite{wang2019edvr,chan2021basicvsr} would overlook their inter-correlation on motion modeling. 

\begin{figure}[tb]
\centering
\includegraphics[width=8.3cm]{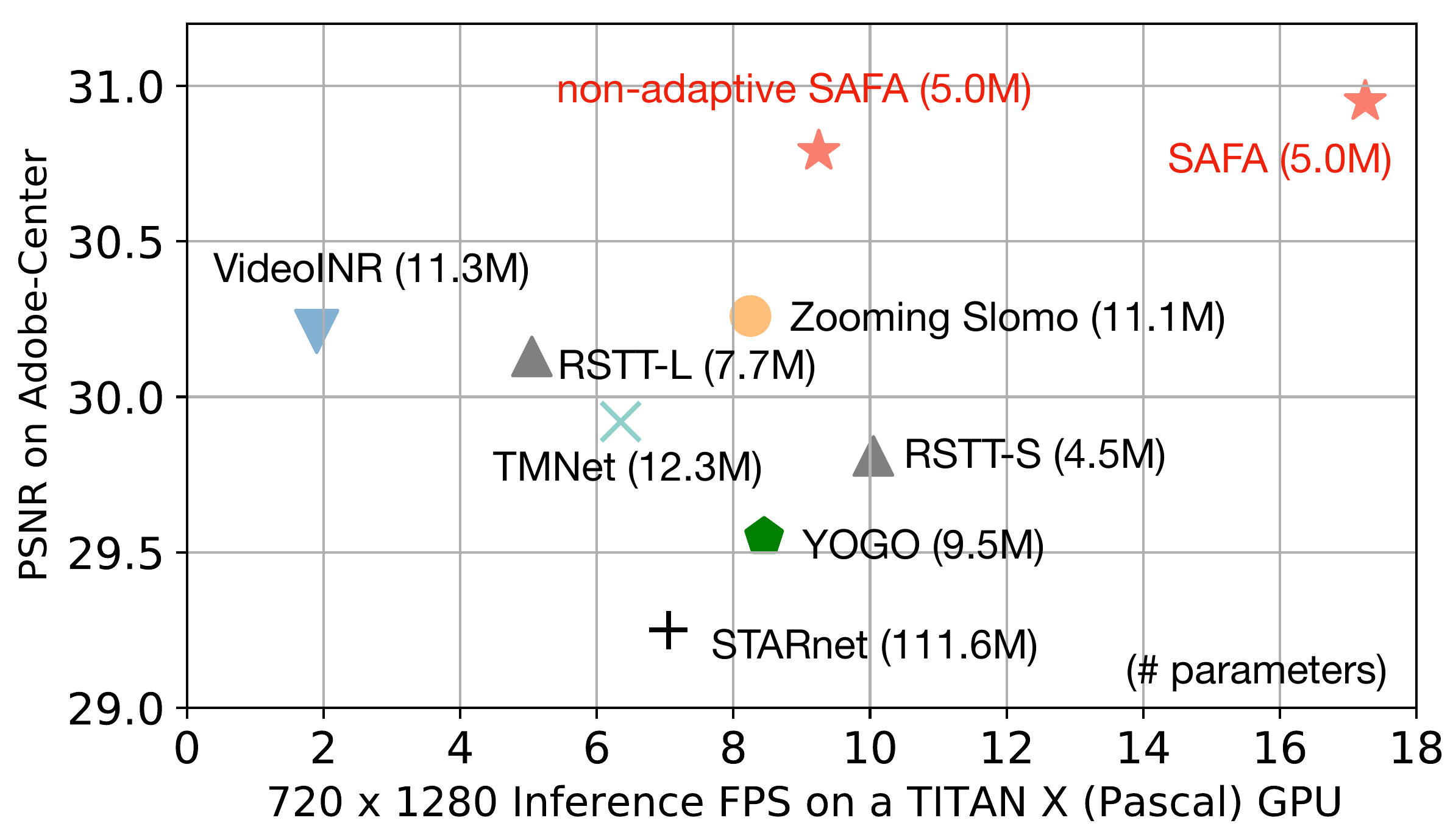}
\caption{\textbf{Performance comparison on the Adobe240-\textit{Center} dataset~\cite{su2017adobe}}. SAFA outperforms the other methods in terms of inference speed and PSNR (dB) metric.
}
\label{fig:performance}
\end{figure}

\begin{figure}[t]
	\centering
	\includegraphics[width=8.3cm]{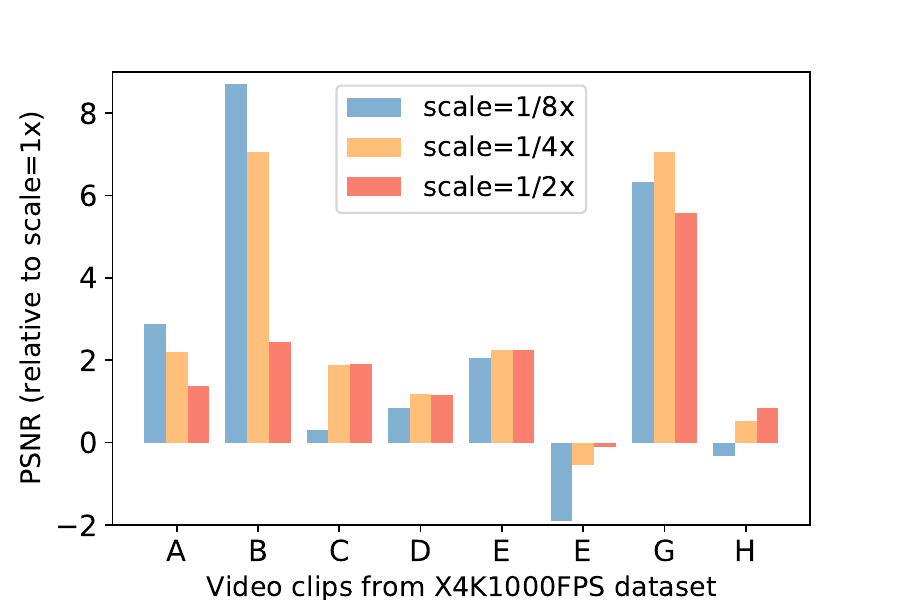}
	\caption{\textbf{$8\times$ VFI using RIFE~\cite{huang2022rife} under different resolution scales of flow computing on the X4K1000FPS dataset~\cite{sim2021xvfi}}. Different video clips have respective suitable processing scales. }
	\label{fig:xvfi}
\end{figure}

Recent one-stage STVSR methods~\cite{xiang2020zooming,xu2021tmnet,chen2022videoinr} have made great progress. They model temporal motion~\cite{huang2022rife} by incorporating motion representation operators, \eg, deformable convolution~\cite{dai2017deformable} and DConvLSTM~\cite{xiang2020zooming}, into the convolutional neural network~(CNN) backbones. Integrating additional components into these STVSR models would increase their complexity and overhead, which is not conducive to fulfilling the flexible requirements of diverse scenarios. 
Therefore, the complex structures adopted by previous methods might impede the progress of future STVSR methods. Motion estimation components have become more and more complex to manage various scenes and resolutions. Many models are trained on small (\eg, $256\times256$) patches~\cite{xue2019video}, yet relying on hand-crafted multi-stage and multi-scale strategies~\cite{niklaus2020softmax,kong2022ifrnet} to process high-resolution videos~\cite{sim2021xvfi,park2021ABME}. Even with these efforts, current method still perform considerably differently as the processing resolution changes and per video. We show the results of a toy experiment in Figure~\ref{fig:xvfi} and put the specific analysis in~\S\ref{sec:design}.


To mitigate the issue of excessive model complexity, it is imperative to develop efficient and streamlined one-stage STVSR models.
As suggested by~\cite{chan2021basicvsr,radosavovic2020designing}, we decouple various functional components and integrate promising techniques from high-level vision. 
Since the input frames of STVSR are in low resolution, we focus on aggregating contextual features to supplement the lack of information in pixel-wise alignment~\cite{niklaus2020softmax,huang2022rife,kong2022ifrnet}.
To improve motion estimation and model efficiency, we propose a Scale-Adaptive Flow Estimation~(SAFE) block to construct our model, Scale-Adaptive Feature Aggregation~(SAFA). 

Firstly, the flow estimation blocks recurrently update an intermediate hidden state that encodes motion information based on the contextual feature. %
Then, we use the dynamic routing technique~\cite{elsken2019neural} to adapt SAFA to diverse moving objects and motion scales. Specifically, we introduce a data-dependent scale selector using a Bernoulli distribution, enabling SAFE to well utilize the scalable advantage of motion estimation. The multi-branch selection reduces the computational overhead and helps SAFA fit the data more accurately. 
The propagation of image information and contextual features complement each other for better STVSR performance.
Experiments on these modules further validate that SAFA is a simple yet effective method that outperforms state-of-the-art STVSR methods.
As shown in Figure~\ref{fig:performance}, SAFA is faster and achieves over 0.5dB higher PSNR results than the comparison STVSR methods like VideoINR~\cite{chen2022videoinr}, TMNet~\cite{xu2021tmnet}, and RSTT~\cite{geng2022rstt} on the Adobe240~\cite{su2017adobe} dataset.

In summary, our contributions are three-fold:
\vspace{-0.4em}
\begin{itemize}
    \item We introduce an innovative SAFE block that estimates motion in an iterative manner with trainable block-wise scale selection.
    \item With SAFE block, we propose an efficient SAFA method for STVSR. 
    The modular design of SAFA makes it a simple yet efficient solution for STVSR with input-adaptive inference customization.
    \item Experiments show that SAFA quantitatively and qualitatively outperforms state-of-the-art STVSR methods, while exhibiting faster inference speed and fewer parameter amounts.
\end{itemize}
\vspace{-0.4em}

	\begin{figure*}[tb]
	\centering
	\includegraphics[width=17cm]{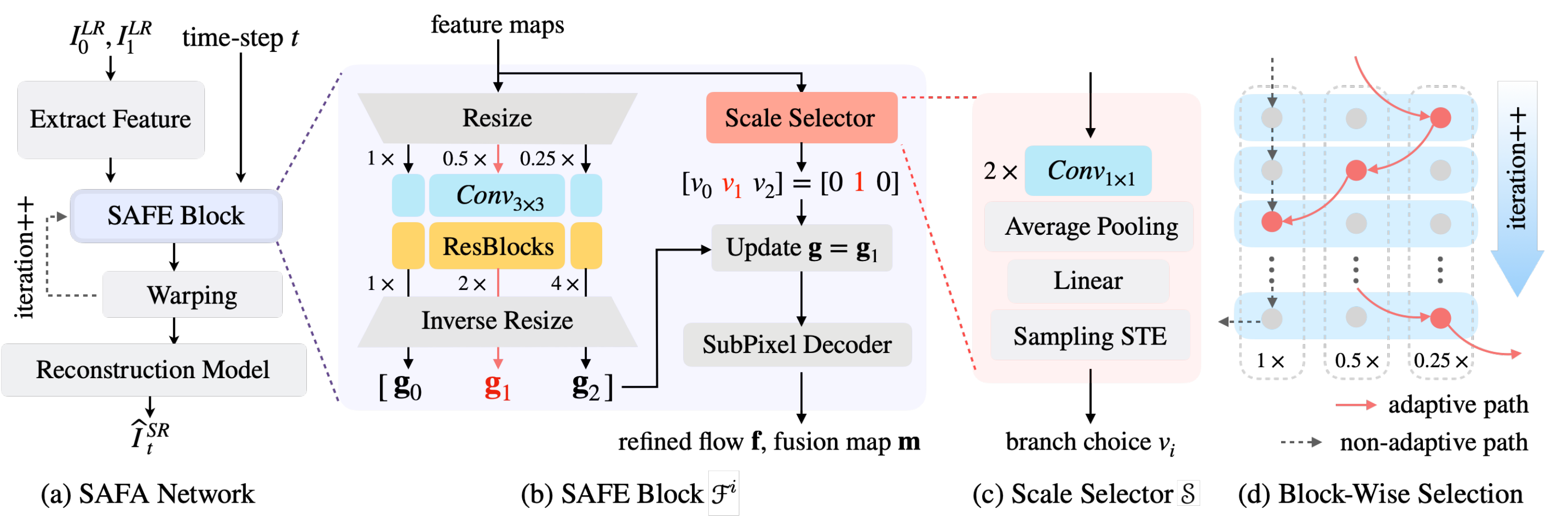}
	\caption{\textbf{Architecture of our SAFA}. (a) We design SAFA with low intra-module design complexity. The motion is iteratively estimated using SAFE blocks. (b) A SAFE block consists of several neural layers. We warp the image feature and update the hidden state \textbf{g}. We use a scale selector $\mathcal{S}$ to choose a suitable processing scale. (c) The scale selector is a small neural network with a differentiable sampling STE~\cite{bengio2013estimating,dorefanet}. (d) The adaptive selection of the sub-network is block-wise. }
	\label{fig:main}
 \vspace{-0.5em}
\end{figure*}

\section{Related Work}

\subsection{Optical Flow Estimation} 
Optical flow estimation has garnered increasing attention in recent research.
Pioneering works such as FlowNet~\cite{dosovitskiy2015flownet} and FlowNet2.0~\cite{ilg2017flownet} have paved the way by learning to estimate dense optical flow fields from synthetic datasets.
Efficient networks like PWC-Net~\cite{sun2018pwc} and LiteFlowNet~\cite{hui2018liteflownet} have been constructed by warping features and computing cost volume at various pyramid levels, thereby calculating optical flow in a coarse-to-fine manner~\cite{ranjan2017optical,hu2022restore}.
RAFT~\cite{teed2020raft} has achieved a significant breakthrough by iteratively updating a flow field through a recurrent ConvGRU~\cite{shi2015convolutional}.
The iterative framework introduced by RAFT has further evolved into new variants~\cite{jiang2021learning,sui2022craft}.
Transformer-based flow estimation models such as GMFlow~\cite{xu2022gmflow}, FlowFormer~\cite{huang2022flowformer}, and VideoFlow~\cite{shi2023videoflow} have also shown promising advancements in the field.
Moreover, several studies have successfully generalized optical flow techniques for downstream tasks~\cite{xue2019video,han2022realflow}.
In this work, we introduce scale-adaptive path selection into task-oriented flow-based propagation for efficient STVSR.

\subsection{Video Super-Resolution}
In recent years, research on VSR has gradually shifted from sliding-window framework~\cite{xue2019video,wang2019edvr,tian2020tdan} to recurrent framework~\cite{huang2017video,fuoli2019efficient,isobe2020video,chan2021basicvsr}. The recurrent framework can leverage long-term motion information for better performance. The deformable-model-based alignment scheme~\cite{wang2019edvr,chan2021understanding} is an extension of flow-based alignment and is widely used. Benefiting from in-depth research on long-term feature alignment and propagation, the work of BasicVSR~\cite{chan2021basicvsr,chan2022basicvsrpp} has achieved very competitive performance. We refer readers to~\cite{chan2022basicvsrpp} for detailed technologies of long-term motion modeling. There are also some explorations of reusing optical flow between different frames~\cite{zhang2022optical}.
These techniques are orthogonal to ours, as we focus more on streaming video processing schemes that do not rely on many distant frames.

\subsection{Video Frame Interpolation} 

Most existing VFI methods can be divided into two main paradigms: kernel-based methods~\cite{niklaus2017video,Niklaus_ICCV_2017} and flow-based methods~\cite{liu2017video,jiang2018super,zhang2020flexible,park2021asymmetric,huang2022rife,kong2022ifrnet}. They share the idea that the pixels of the generated frames come from nearby regions of the adjacent frames.
The kernel-based methods, \eg,
AdaConv~\cite{niklaus2017video} and SepConv~\cite{Niklaus_ICCV_2017}, implicitly model the motion by estimating spatially adaptive kernels for different output pixels.
On the other hand, flow-based methods explicitly model the motion of objects. Specifically, DVF~\cite{liu2017video} predicts the intermediate flow and occlusion mask jointly using CNN. SuperSloMo~\cite{jiang2018super} uses two U-Net~\cite{ronneberger2015u} models to estimate the bi-directional flows and occlusion mask successively. The research on optical flow is gradually deepening. ABME~\cite{park2021asymmetric} explores asymmetric motion estimation. RIFE~\cite{huang2022rife} and IFRNet~\cite{kong2022ifrnet} propose to build more refined models and learn intermediate flow estimation via knowledge distillation.
AdaCoF~\cite{lee2020adacof} and EDSC~\cite{cheng2020multiple} explore the integration of both paradigms under the multi-flow collaboration framework.
Some recent works focus on specific scenarios, such as 
animations~\cite{siyao2021deep,chen2021improving} and near-duplicate photos~\cite{reda2022film}.
Despite that VFI models are becoming more sophisticated, such as VFIformer~\cite{lu2022video} and EMA~\cite{zhang2023extracting}, 
we notice that the critical path is still the feature extraction and motion estimation. With this clue, our designed model is modular with efficient flow-based feature aggregation.

\subsection{Space-Time Video Super-Resolution} 
Standing on the early pioneer explorations~\cite{shechtman2005space,mudenagudi2010space,shahar2011space}, STVSR has achieved new developments in the era of deep learning. FISR~\cite{kim2020fisr} is among the pioneering work to unify VFI and VSR networks. As concurrent work, Xiang~\etal~\cite{xiang2020zooming} proposed to aggregate temporal information by a unified deformable ConvLSTM~\cite{dai2017deformable}. Haris~\etal~\cite{haris2020space} proposed the STARnet to leverage mutually informative relationships between temporal and spatial dimensions, and fuse contexts at different resolution scales.
To further improve STVSR performance, the work of STVUN~\cite{kang2020deep}, MBNet~\cite{zhou2021video}, and YOGO~\cite{hu2022you} exploit the inter-correlation, interaction, and integration of propagation schemes between VSR and VFI tasks, respectively.
Besides, TMNet~\cite{xu2021tmnet} modulates the deformable convolution kernels for arbitrary-time frame interpolation.
CycMu-Net~\cite{hu2022spatial} makes full use of spatial-temporal correlations via mutual learning between two processing dimensions.
VideoINR~\cite{chen2022videoinr} utilizes implicit neural representation for the arbitrary scale of temporal and spatial super-resolution. Our work focuses on improving the overall performance of STVSR through explicit and scale-adaptive feature aggregation.
	\section{Proposed Method}

Here, we first overview SAFA for STVSR in~\S\ref{sec:overview}.
Then, we describe the scale inconsistency issue and scale-adaptive flow estimation in~\S\ref{sec:design}.
Finally, we provide the training details in \S\ref{sec:implement} for reproducible research.


\subsection{Overall Network}
\label{sec:overview}


%
Following previous settings~\cite{xiang2020zooming,xu2021tmnet}, we first use bicubic interpolation to upsample the input LR frames $\{I_0^{LR}, I_1^{LR}\}$, by a scaling factor of $4\times$.
Here, we still keep the index ``${LR}$'' on the upsampled frames as $\{I_0^{LR}, I_1^{LR}\}$ for descriptive convenience.
Thus, in SAFA, the input and output frames are of the same size.
As shown in Figure~\ref{fig:main}, the proposed SAFA basically contains three parts: a spatial feature extractor $f_{\theta}$, our proposed SAFE network $\mathcal{F}$ for temporal feature aggregation, and a reconstruction module $\mathcal{R}$.

SAFA network firstly extracts the feature maps $\textbf{c}_0$ and $\textbf{c}_1$ of the two adjacent frames $I_0^{LR}$ and $I_1^{LR}$, respectively, by the feature extractor $f_{\theta}$. That is, $f_{\theta}(I_0^{LR})=\textbf{c}_0$, $f_{\theta}(I_1^{LR})=\textbf{c}_1$. The extracted feature maps will be employed to calculate the intermediate flow field and further propagated to fuse the latent feature of the frame at a target intermediate moment $t$. In the era of deep learning, there are many milestone works extracting useful representations from images~\cite{krizhevsky2017imagenet,simonyan2014very,he2016deep,liang2021swinir}. Inspired by the successful FPN networks~\cite{lin2017feature,chen2021you}, we employ the ResNet-18~\cite{he2016deep} as our feature extractor $f_{\theta}$ and integrate the feature maps output by the input stem, stage-1, and stage-2. To seamlessly integrate these feature maps of varying sizes, we use a $1\times1$ convolutional layer and bilinear upsampling to adjust the channel numbers and spatial sizes, respectively, illustrated in \textbf{Appendix}.
\begin{figure}[t]
	\centering
	\includegraphics[width=8cm]{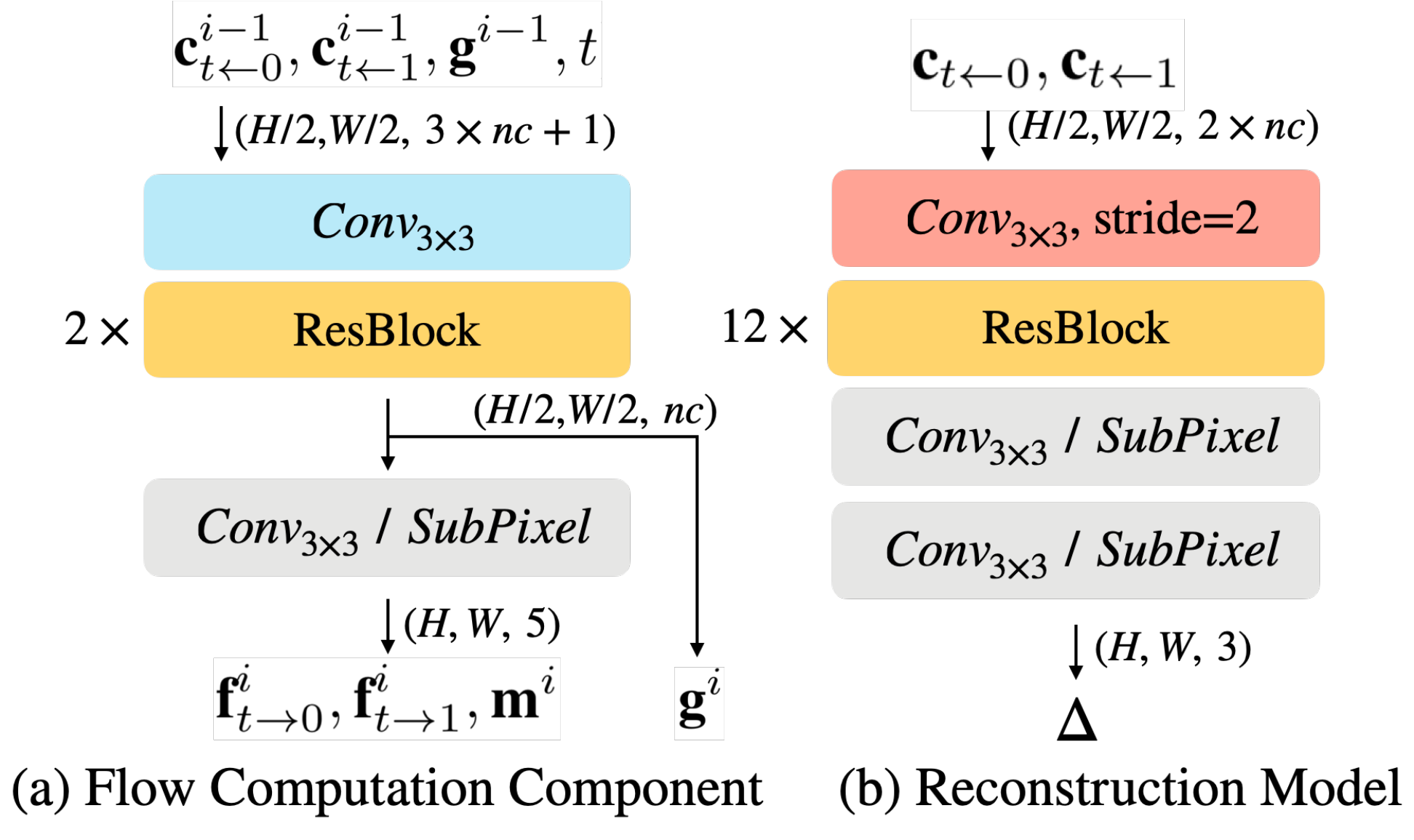}
	\caption{\textbf{Architecture of a single-path SAFE block $\mathcal{F}^{i}$ (a) and reconstruction model $\mathcal{R}$~(b)}. Multiple inputs of each module are combined with concatenate operator. The hidden layers use PReLU~\cite{he2015delving} as nonlinear activation. $nc$ is a hyper-parameter that controls the complexity of SAFA.} 
	\label{fig:model}
\end{figure}

With the extracted feature maps $\{\textbf{c}_0, \textbf{c}_1\}$, we then perform temporal feature aggregation by SAFE blocks.
The SAFE network $\mathcal{F}$ estimates the bidirectional intermediate flow fields $\textbf{f}_{t\rightarrow 0}$ and $\textbf{f}_{t\rightarrow 1}$, as well as a fusion map $\textbf{m}\in[0,1]$ to model object occlusion~\cite{liu2017video,jiang2018super}.
That is,
\begin{equation}
(\textbf{f}_{t\rightarrow 0},\textbf{f}_{t\rightarrow 1},\textbf{m})=\mathcal{F}(\textbf{c}_0,\textbf{c}_1,t).
\end{equation}
To synthesize accurate target frame at moment $t$, we further propagate the LR frames $\{I_0^{LR},I_1^{LR}\}$ and the feature maps $\{\textbf{c}_0,\textbf{c}_1\}$ to those at the target intermediate moment $t$.
This is implemented by pixel-wise backward warping $\backwardwarp$ with the flow fields $\{\textbf{f}_{t\rightarrow0},\textbf{f}_{t\rightarrow 1}\}$, as follows:
\begin{equation}
\widehat{I}_{t\leftarrow 0} = \backwardwarp(I_0^{LR}, \textbf{f}_{t\rightarrow 0}),\quad \widehat{I}_{t\leftarrow 1} = \backwardwarp(I_1^{LR}, \textbf{f}_{t\rightarrow 1}),
\end{equation}
\begin{equation}
\textbf{c}_{t\leftarrow 0} = \backwardwarp(\textbf{c}_0, \textbf{f}_{t\rightarrow 0}),\quad \textbf{c}_{t\leftarrow 1} = \backwardwarp(\textbf{c}_1, \textbf{f}_{t\rightarrow 1}),
\end{equation}
After obtaining the propagated frames $\{\widehat{I}_{t\leftarrow0},\widehat{I}_{t\leftarrow1}\}$ and feature maps $\{\textbf{c}_{t\leftarrow0},\textbf{c}_{t\leftarrow1}\}$, we estimate a residual refinement $\Delta$ of the SR frame $\widehat{I}_t^{SR}$ by the reconstruction module:
\begin{equation}
\Delta = \mathcal{R}(\textbf{c}_{t\leftarrow 0}, \textbf{c}_{t\leftarrow 1}).
\end{equation}
Finally, the reconstructed frame $\widehat{I}_t^{SR}$ is obtained by
\begin{equation}
\widehat{I}_t^{SR} = [\textbf{m}\odot\widehat{I}_{t\leftarrow0} + (1 - \textbf{m})\odot\widehat{I}_{t\leftarrow1}] + \Delta,
\end{equation}
where $\odot$ means pixel-wise multiplication.

The obtained target frame $\widehat{I}_t^{SR}$ is consisted of two parts: the pixel-wise fusion of the predicted intermediate frames $\{\widehat{I}_{t\leftarrow0},\widehat{I}_{t\leftarrow1}\}$ propagated from the input LR images~\cite{zhou2016view,liu2017video} and the residual refinement $\Delta$.
$\textbf{f}$, $\textbf{m}$ and $\Delta$ are of the same sizes with the HR frame.
We have $0\leq\textbf{m}\leq1$ and $-1\leq\Delta\leq1$, since $\textbf{m}$ and $\Delta$ are obtained by the Sigmoid and Tanh functions, respectively.
The feature maps $\textbf{c}_0$ and $\textbf{c}_1$ are at $1/2$ resolution of the input frames.
When warping the feature maps $\{\textbf{c}_0,\textbf{c}_1\}$, we need to resize the flow fields bilinearly to match their resolution.
The reconstruction module $~\mathcal{R}$ is mainly consisted of ResBlocks~\cite{he2016deep} and SubPixel~\cite{shi2016real} operator, as shown in Figure~\ref{fig:model}.



\subsection{Scale-Adaptive Flow Estimation}
\label{sec:design}

\noindent The versatile motion amplitudes and object sizes bring great challenges into flow estimation~\cite{niklaus2020softmax,teed2020raft,sim2021xvfi}. In high-level vision tasks like image classification, researchers often enlarge the size of input images to obtain more accurate prediction results~\cite{Tan2019EfficientNetRM}. However, the effective receptive fields of flow-based video synthesizing models~\cite{huang2022rife} may not fully cover the objects of interests and their \textbf{large} motions in HR frames. We not only want the spatial information provided by the HR frame, but also want the receptive field of the model to cover the large motion of the object. This inspires us to design dynamic inference paths in the model to handle different scenarios.

\noindent\textbf{Scale inconsistency issue}. We clarify the scale inconsistency issue by a toy experiment. We use a pre-trained VFI model, RIFE~\cite{huang2022rife}, to perform $8\times$ time scale VFI on the X4K1000FPS~\cite{sim2021xvfi} benchmark. The details of the evaluation settings are consistent with those in~\cite{sim2021xvfi}. We use RIFE~\cite{huang2022rife} to infer the flow fields at different scales and then interpolate the intermediate frames. As shown in Figure~\ref{fig:xvfi}, each video clip gains improvement on PSNR with different inference scales, even if RIFE~\cite{huang2022rife} has equipped the coarse-to-fine flow estimation~\cite{hu2016efficient,sun2018pwc}. By default, RIFE selects the $1/4\times$ scale and achieves an average PSNR of $30.58$dB. It further gains an improvement of about $0.4$dB if we manually pick the inference scales with best results for each video clip. This demonstrates that it is useful to select an adaptive inference scale for each video clip. However, it is laborious to manual inspection and selection of the inference scale clip by clip.
%


\noindent
\textbf{Intermediate flow estimation}.
We follow the trial-and-error manner for flow estimation~\cite{ilg2017flownet,sun2018pwc,hur2019iterative,teed2020raft}. That is, the model estimates an optical flow field and uses the current field to warp the spatial information of the original input frame or feature map, and so on. We encode the time-step $t$ as a separate channel~\cite{huang2022rife,kong2022ifrnet}, and feed it into the flow estimation module along with the image features. We employ a bidirectional network to project the features $\{\textbf{c}_0, \textbf{c}_1\}$ from both directions to that at the moment $t$, under an iterative trial-and-error manner. Formally, we denote the $i^{th}$ SAFE block as $\mathcal{F}^i$, as shown in Figure~\ref{fig:model}. They share the same model structure, but with independent parameters:
\begin{equation}
\mathcal{F} = \{\mathcal{F}^1, \mathcal{F}^2,..., \mathcal{F}^K\}.
\end{equation}
Here, we set $K=6$ in SAFA.

As shown in Figure~\ref{fig:main}, we iteratively update the hidden state $\textbf{g}$ that encodes motion. In each iteration, we use the current estimated flow field to propagate extracted feature $c$ to obtain the bidirectional feature maps $\{\textbf{c}_{t\leftarrow0}^{i-1},\textbf{c}_{t\leftarrow1}^{i-1}\}$ of the intermediate frame at a target moment $t$. Then these feature maps are input to the next SAFE block. The calculation process of $K$ iterations is as follows:
\begin{equation}
\textbf{g}^{0} = \text{Conv}_{3\times 3}(\text{Concat}(\textbf{c}_0, \textbf{c}_1)),
\end{equation}
\begin{equation}
\textbf{g}^{i} = \mathcal{F}^i(\textbf{c}_{t\leftarrow 0}^{i-1}, \textbf{c}_{t\leftarrow 1}^{i-1}, \textbf{g}^{i-1}, t),\ i=1,2,...,K.
\end{equation}
Finally, we get iteratively refined flow fields $\{\textbf{f}_{t\rightarrow 0},\textbf{f}_{t\rightarrow 1}\}$ and an occlusion mask map $\textbf{m}$, which will be used to propagate spatial information of input frames and their feature maps, as described in~\S\ref{sec:overview}.


\noindent
\textbf{Input-adaptive scale selection}.\ To endow SAFA with the ability of dynamic scale selection, we propose a SAFE block to update the hidden states
$\textbf{g}=\mathcal{F}(x)$ on different scales, and select the one with the best performance, as shown in Figure~\ref{fig:main}. The computational costs for calculating the state $\textbf{g}^{1/s}$ with $1/s$ resolution are $1/s^2$ of that in the original resolution. This provides SAFA an appropriate path to predict the intermediate frame at a target moment $t$ from two given frames.

We introduce the STE sampling~\cite{bengio2013estimating,dorefanet,hu2023dmvfn} for multi-branch selection.
The neural network of scale selector $\mathcal{S}$  estimates the selection probability $P_i$ of $B$ branches, where $i=0, 1, ..., B-1$ and $0\leq P_i \leq 1$. We then choose a branch based on $P_i$: 
\begin{equation}
\textbf{Forward}: p_i\sim Bernoulli(P_i),\ p_i\in\{0, 1\},
\end{equation}
\begin{equation}
v_0 = p_0,\ v_1 = (1 - p_0) * p_1, 
\end{equation}
\begin{equation}
v_2 = (1 - p_0) * (1-p_1) * p_2,\ ..., 
\end{equation}
\begin{equation}
\textbf{g} = \sum_{i=0}^{B-1} v_i * \textbf{g}_i,
\end{equation}
\begin{equation}
\textbf{Backward}: \frac{\partial o}{\partial {P_i}} = \frac{\partial o}{\partial p_i},
\end{equation}
where we calculate $\textbf{g}$ in multiple branches gated by the branch choice $v_i$, $o$ is the objective function. Essentially, the above formula performs multiple dual-branch selections to construct a computable multi-branch selection.

In practice, we use three branches of different scales $\{\textbf{g}^{1}, \textbf{g}^{1/2}, \textbf{g}^{1/4}\}$, \ie, $B=3$. During training, our multi-branch design increases the training overhead by $30\%$ compared to that with a single branch $\textbf{g}^1$. In the inference stage, only one branch needs to be calculated. The parameters of the different branches here are shared. On the Adobe240 benchmark~\cite{su2017adobe}, the runtime of SAFA is 60\% of non-adaptive SAFA.
The scale selector $\mathcal{S}$ occupies only 2\% of computation overhead by SAFA model.
\subsection{Implementation Details}
\label{sec:implement}


SAFA model is optimized from scratch~(random initialization) by Adam~\cite{adam}. We train SAFA end-to-end using $L_1$ loss function with $32\times32$ down-sampled patches for $600,000$ iterations. The batch size in training is $24$. We gradually reduce the learning rate from $2\times10^{-4}$ to $0$ using cosine annealing scheme~\cite{xiang2020zooming,xu2021tmnet} during the training process. We use a sliding window with a length of $9$ to select frames from videos. In one training sample, we select the first and last frames of the sliding window as the inputs and the intermediate frames are taken as the ground truths. The corresponding time-step $t$ is from $\{0, \frac{1}{8}, \frac{2}{8}, ...,1\}$.  Note that we randomly select one intermediate frame for STVSR prediction during the training stage. We randomly augment the training sample by horizontal and vertical flipping, rotating~($90^{\circ}$, $180^{\circ}$ and $270^{\circ}$), temporal order reversing. The training is performed on four Pascal TITAN X GPUs, which takes about $50$ hours. For comparison, VideoINR~\cite{chen2022videoinr} and TMNet~\cite{xu2021tmnet} require more than one week for training.

	\begin{table*}[t]
\caption{\textbf{Quantitative comparison} on the Vid4~\cite{liu2011vid4}, GoPro~\cite{nah2017gopro}, and Adobe240~\cite{su2017adobe} datasets. We omit some results of methods that can only synthesize frames at fixed times $t=0.5$. Some of the previous methods are reported by VideoINR~\cite{chen2022videoinr}.}
\resizebox{\textwidth}{!}{
\begin{tabular}{cc|ccccccccccc}
\hline
\multirow{2}{*}{\begin{tabular}[c]{@{}c@{}}
VFI\\Method\end{tabular}} & \multirow{2}{*}{\begin{tabular}[c]{@{}c@{}}VSR\\Method\end{tabular}} & \multicolumn{2}{c}{Vid4} & \multicolumn{2}{c}{GoPro-\textit{Center}} & \multicolumn{2}{c}{GoPro-\textit{Average}} & \multicolumn{2}{c}{Adobe-\textit{Center}} & \multicolumn{2}{c}{Adobe-\textit{Average}} & \multirow{2}{*}{\begin{tabular}[c]{@{}c@{}} \# Param\\   (M)\end{tabular}} \\ 
 & & PSNR & SSIM & PSNR & SSIM & PSNR & SSIM & PSNR & SSIM & PSNR & SSIM & \\ 
\hline \hline
DAIN~\cite{bao2019depth} & EDVR~\cite{wang2019edvr} & 23.48 & 0.6547 & 28.01 & 0.8239 & 26.37 & 0.7964 & 27.06 & 0.7895 & 26.01 & 0.7703 & 44.7 \\ 
DAIN~\cite{bao2019depth} & BasicVSR~\cite{chan2021basicvsr} & 23.43 & 0.6514 & 28.00 & 0.8227 & {26.46} & 0.7966 & 27.07 & 0.7890 & {26.23} & {0.7725} & 30.3 \\ 
IFRNet~\cite{kong2022ifrnet} & EDVR~\cite{wang2019edvr} & 23.68 & 0.6515 & 28.49 & 0.8379 & 26.41 & 0.7980 & 27.31 & 0.7981 & 26.12 & 0.7710 & 25.7 \\ 
IFRNet~\cite{kong2022ifrnet} & BasicVSR~\cite{chan2021basicvsr} & 23.76 & 0.6603 & 28.55 & 0.8392 & 26.58 & 0.8012 & 27.44 & 0.8005 & 26.35 & 0.7769 & 11.3 \\
\hline
\multicolumn{2}{c|}{Zooming SloMo~\cite{xiang2020zooming}} & {25.72} & {0.7717} & {30.69} & {0.8847} & - & - & {30.26} & {0.8821} & - & - & {11.1} \\ 
\multicolumn{2}{c|}{TMNet~\cite{xu2021tmnet}} & \underline{25.96} & \underline{0.7803} & 30.14 & 0.8692 & {28.83} & {0.8514} & 29.41 & 0.8524 & {28.30} & {0.8354} & {12.3} \\ 
\multicolumn{2}{c|}
{RSTT-L~\cite{geng2022rstt}} & 25.94 & 0.7801 & 30.37 & 0.8762 & - & - & 30.13 & 0.8754 & - & - & {7.7} \\ 
\multicolumn{2}{c|}{VideoINR-\textit{fixed}~\cite{chen2022videoinr}} & {25.78} & {0.7730} & {30.73} & {0.8850} & - & - & {30.21} & {0.8805} & - & - & {11.3} \\ 
    \multicolumn{2}{c|}{VideoINR~\cite{chen2022videoinr}} & 25.61 & 0.7709 & {30.26} & {0.8792} & {29.41} & {0.8669} & {29.92} & {0.8746} & {29.27} & {0.8651} & {11.3} \\ 
    \multicolumn{2}{c|}{non-adaptive SAFA} & 25.82 & 0.7759 & \underline{30.97} & \underline{0.8871} & \underline{30.01} & \underline{0.8719} & \underline{30.79} & \underline{0.8854} & \underline{29.92} & \underline{0.8736} & \textbf{5.0} \\
    \multicolumn{2}{c|}{SAFA} & \textbf{25.98} & \textbf{0.7807} & \textbf{31.28} & \textbf{0.8894} & \textbf{30.22} & \textbf{0.8761} & \textbf{30.97} & \textbf{0.8878} & \textbf{30.13} & \textbf{0.8782} & \textbf{5.0} \\ 
    
\hline
\end{tabular}
}
\label{tab:sota-result}
\end{table*}

\section{Experiments}
\definecolor{mygray}{gray}{.9}
\subsection{Datasets and Evaluation Metrics} 
For STVSR methods, a typical training or testing sample of video frame tuples is $\{I_0^{LR}, I_t^{HR}, I_1^{LR}, t\}$. Here, the input LR frames $I_0^{LR}$ and $I_1^{LR}$ are bicubicly down-sampled by $0.25\times$ from the corresponding HR frames $I_0^{HR}$ and $I_1^{HR}$, respectively. Following VideoINR~\cite{chen2022videoinr}, we train the comparison methods on the Adobe240 training dataset~\cite{su2017adobe}, which contains $100$ 720p videos and each video has about $3,000$ frames. Our evaluation is on the following datsets:

\noindent
\textbf{Vid4}~\cite{liu2013bayesian} is a popular test set containing 171 frames of 480p. On this dataset, we conduct STVSR experiments for single-frame interpolation. Following VideoINR~\cite{chen2022videoinr}, the metrics reported on Vid4 differ from some previous literatures~\cite{xiang2020zooming,xu2021tmnet} due to using a smaller sliding window. 

\noindent
\textbf{Adobe240}~\cite{su2017adobe} contains 17 videos in its test subset. We take the first frame out of every eight frames (\ie, $1^{st}$, $9^{th}$, $17^{th}$ frame,...) to produce the input LR frames, and use them to interpolate the intermediate frames.

\noindent{\textbf{GoPro}}~\cite{nah2017gopro} contains $11$ 720p and 240FPS videos. These videos are mostly street scenes captured by action cameras. We pre-process these videos in the same way as the Adobe240~\cite{su2017adobe} test set. We do not use the training set~\cite{nah2017gopro}.

\noindent\textbf{X4K1000FPS} is a high frame rate 4K dataset~\cite{sim2021xvfi} containing $15$ street scenes for testing.

For Adobe240~\cite{su2017adobe} and GoPro~\cite{nah2017gopro} datasets, we separately evaluate the average metrics of the center (\ie, $1^{st}$, $5^{th}$, $9^{th}$, ...) frames and all output frames, which are denoted as \textit{-Center} and \textit{-Average}, respectively. We use the metrics of Peak Signal-to-Noise Ratio (PSNR) and structural similarity (SSIM)~\cite{wang2004image} for quantitative evaluation. Following previous works~\cite{xiang2020zooming,xu2021tmnet,chen2022videoinr}, the generated images will be evaluated on Y channel of YCbCr space. All the methods are tested on a Pascal TITAN X GPU. To report the runtime, we calculate the average process time for 100 runs after a warm-up process.

\subsection{Comparisons to State-of-the-Arts}

\noindent
\textbf{Comparison methods}.
We compare SAFA with six two-stage  ``VFI+VSR'' methods and four one-stage \sota~STVSR methods~\cite{xiang2020zooming, xu2021tmnet, chen2022videoinr, geng2022rstt}. We reproduce the other methods following VideoINR~\cite{chen2022videoinr}. For two-stage methods, we employ DAIN~\cite{bao2019depth} and IFRNet~\cite{kong2022ifrnet} for VFI, and employ Bicubic Interpolation, EDVR~\cite{wang2019edvr}, and BasicVSR~\cite{chan2021basicvsr} for VSR, both trained on the Adobe240 training set~\cite{su2017adobe}. The one-stage VideoINR~\cite{chen2022videoinr}, RSTT~\cite{geng2022rstt}, and Zooming SloMo~\cite{xiang2020zooming} are trained in similar settings with SAFA. With an additional training stage on Vimeo90K~\cite{xue2019video}, TMNet~\cite{xu2021tmnet} is fine-tuned on the Adobe240~\cite{su2017adobe} training set. As suggested in~\cite{chen2022videoinr}, VideoINR has a variant fixing the interpolation time as $t=0.5$, denoted as VideoINR\textit{-fixed}. 

\noindent
\textbf{Quantitative results}. As shown in Table~\ref{tab:sota-result}, the two-stage methods consuming more parameters perform worse than the one-stage methods on STVSR.
This is consistent with previous literature~\cite{xiang2020zooming,son2021ntire,xu2021tmnet} that demonstrated the benefits of one-stage model design for STVSR methods. 

For one-stage methods, SAFA enables arbitrary time-step interpolation and achieves clearly better performance than the comparison methods on Gopro~\cite{nah2017gopro} and Adobe240~\cite{su2017adobe}. Due to small inter-frame object motion, SAFA is only on par with TMNet~\cite{xu2021tmnet} on Vid4~\cite{liu2011vid4}.

\begin{table}[t]
\caption{\textbf{Quantitative comparison for different time scales} on the GoPro dataset~\cite{nah2017gopro}.}
\centering
\begin{tabular}{c|ccc}
\hline
Time Scale &  TMNet~\cite{xu2021tmnet} & VideoINR~\cite{chen2022videoinr} & SAFA\\ 
\hline \hline
6$\times$   & 30.49dB & {30.78dB} & \textbf{31.67dB} \\

8$\times$   & 28.83dB & {29.41dB} & \textbf{30.22dB} \\

12$\times$  & 26.38dB & {27.32dB} & \textbf{27.97dB}\\
16$\times$  & 24.72dB & {25.81dB} & \textbf{26.32dB}\\
\hline
\end{tabular}
\label{tab:ood}
\end{table}

The original \textit{-Average} evaluation regime can be taken as $8\times$ time scale evaluation. To study the ability of different methods on modeling time-step $t$, we also conduct $6\times, 12\times$, and $16\times$ experiments on GoPro~\cite{nah2017gopro}. As shown in Table~\ref{tab:ood}, SAFA consistently achieves substantial advantages on all regimes over TMNet~\cite{xu2021tmnet} and VideoINR~\cite{chen2022videoinr}.

\begin{figure}[t]
	\centering
	\includegraphics[width=8.2cm]{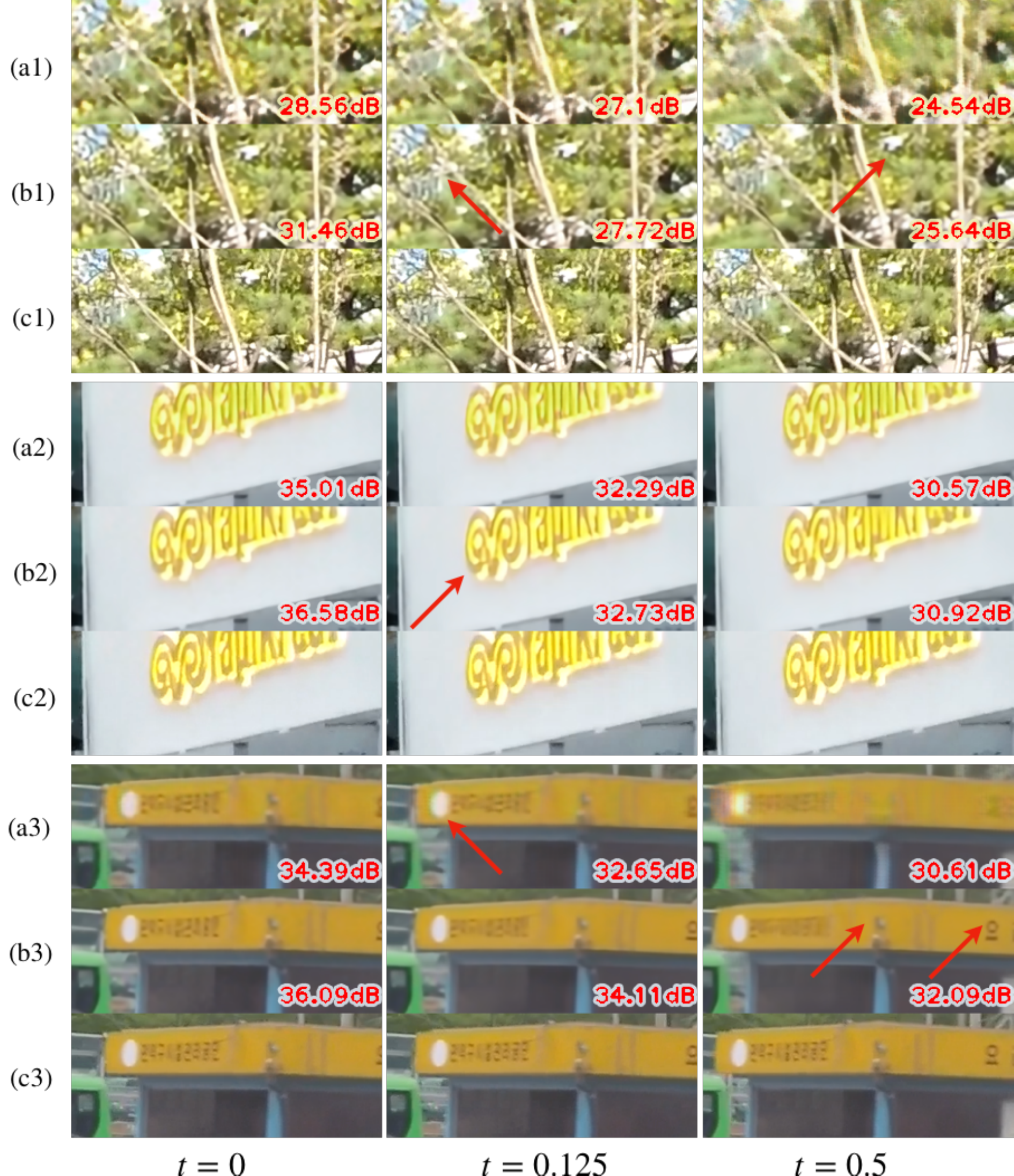}
	\caption{\textbf{Qualitative comparisons on the GoPro dataset~\cite{nah2017gopro}}. (a) VideoINR~\cite{chen2022videoinr}, (b) SAFA, (c) ground truth. We perform $8\times$ time scale interpolation and $4\times$ super-resolution. Three of the eight results are shown. The PSNR annotations are calculated on the full 720p frame. We crop $120\times300$ patches for visualization.}
	\label{gopro}
 \vspace{-1em}
\end{figure}

\begin{figure}[t]
	\centering
	\includegraphics[width=8.3cm]{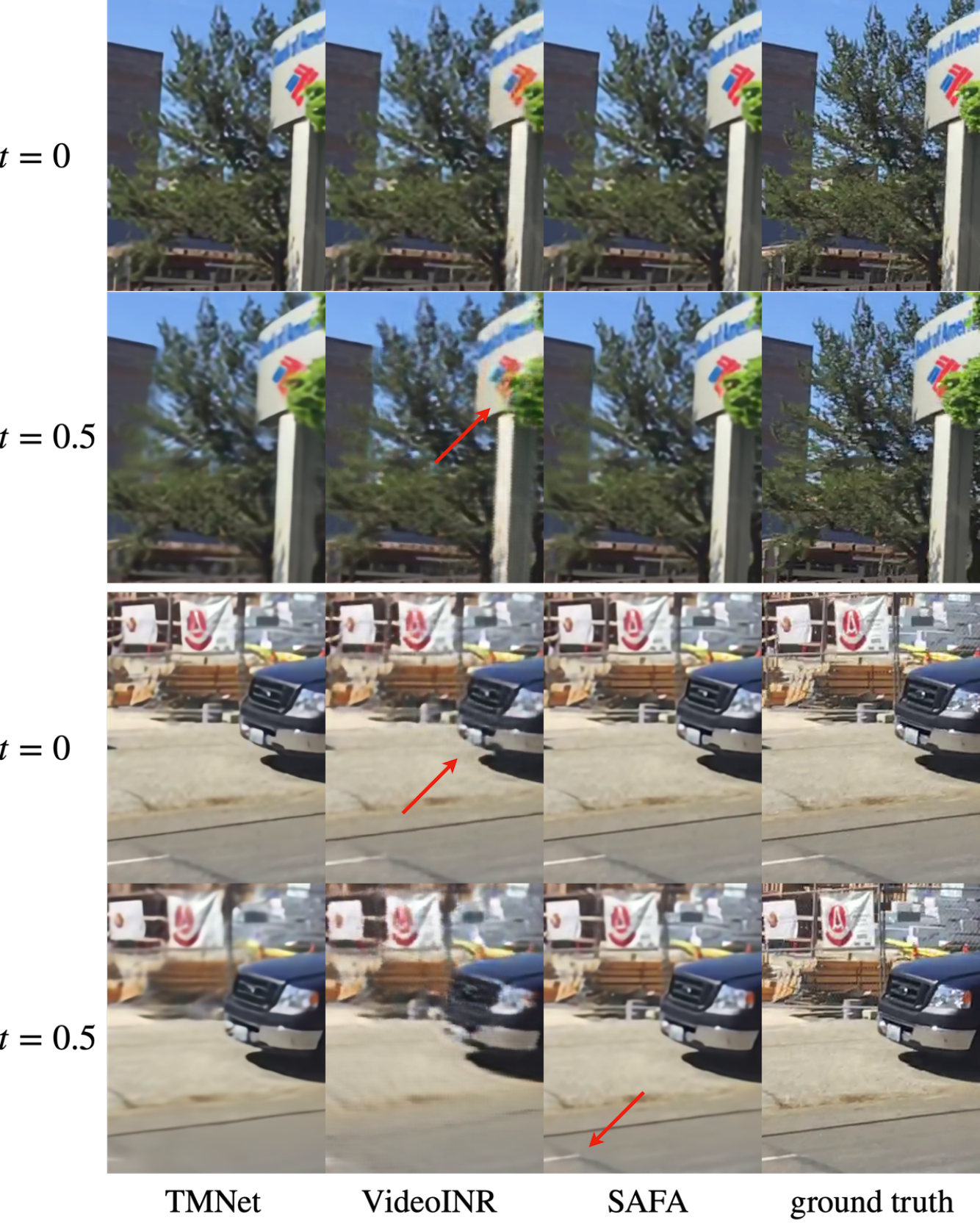}
	\caption{\textbf{Qualitative comparisons on the Adobe240 dataset~\cite{su2017adobe}}. We perform $8\times$ time scale interpolation and $4\times$ super-resolution. Two of the eight results are used for presentation. We crop $320\times240$ patches for visualization.}
	\label{adobe}
 \vspace{-1em}
\end{figure}

\noindent
\textbf{Visual effects}. In Figures~\ref{gopro} and \ref{adobe}, we compare SAFA with TMNet~\cite{xu2021tmnet} and VideoINR~\cite{chen2022videoinr}. SAFA recovers better the details of different scenes, especially on the frames at $t=0.5$. The reconstruction of frames at $t=0.5$ is relatively difficult because it is far from both input frames. We provide some video results in \textsl{Supplementary Materials}.

\begin{figure}[t]
	\centering
	\includegraphics[width=8.2cm]{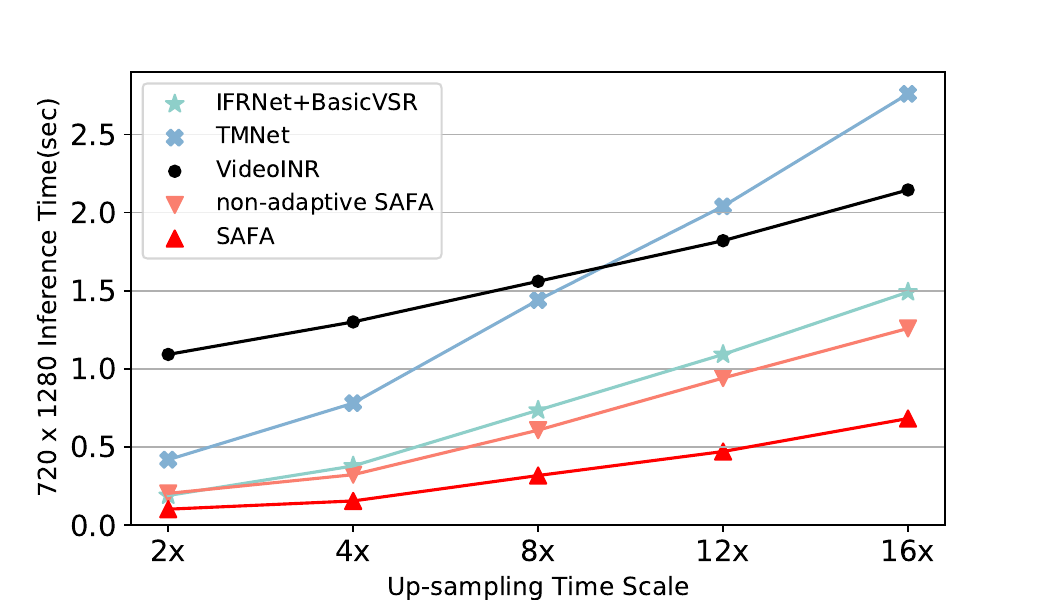}
	\caption{\textbf{Inference time on different up-sampling time scales}. SAFA is efficient and has high accuracy.}
	\label{fig:time}
 \vspace{-1em}
\end{figure}

\noindent
\textbf{Runtime}. We test all comparison methods on the same platform. As shown in Figure~\ref{fig:time}, SAFA has about $0.25\times$ inference time when compared to TMNet~\cite{xu2021tmnet}. The most efficient two-stage method, \ie, IFRNet~\cite{kong2022ifrnet} + BasicVSR~\cite{chan2021basicvsr}, is considerably slower than SAFA.

\subsection{Ablation Study}

\begin{table}[t]
\caption{\textbf{Ablation studies} on the Gopro-\textit{Center}~\cite{nah2017gopro} and Adobe240-\textit{Center}~\cite{su2017adobe} datasets. The default settings of SAFA are marked using gray backgrounds.}
	\resizebox{0.49\textwidth}{!}{\begin{tabular}{lccc}
		\hline
		\multicolumn{1}{c}{\multirow{2}{*}{Setting}} & \multicolumn{1}{c}{GoPro} & \multicolumn{1}{c}{Adobe240} & \multicolumn{1}{c}{\# Param} \\ 
		
		\multicolumn{1}{c}{}                         & \multicolumn{1}{c}{PSNR}   & \multicolumn{1}{c}{PSNR}     & \multicolumn{1}{c}{(M)}            \\ \hline \hline
		\multicolumn{4}{l}{\emph{Feature Extractor $f_{\theta}$}}\\
		\textbf{a1:} \{R18: s2\} & 31.09 & 30.77 &
		5.0\\ 
		{\textbf{a2:} \{R18: s3\}} & 31.01 & 30.68 &
		7.0\\ 
        \rowcolor[rgb]{ .9,  .9,  .9}
		\textbf{a3:} \{R18: stem, s1, s2\} & \textbf{31.28} & \textbf{30.97} &
		5.0\\ 
		\textbf{a4:} \{R50: stem, s1, s2\} & {31.23} & {30.94} &
		{6.0}\\ 
        \textbf{a5:} \{ResNeXt101: stem, s1, s2\} & {31.17} & {30.87} &
		{7.0}\\ \hline
		\multicolumn{4}{l}{\emph{Information Aggregation}}\\
\rowcolor[rgb]{ .9,  .9,  .9}
		\textbf{b1:} image + feature fusion & \textbf{31.28} & \textbf{30.97} & 5.0\\
		\textbf{b2:} image fusion & 30.57 & 30.25 & \textbf{3.3} \\ 
		\textbf{b3:} feature fusion & 31.15 & 30.77 & 5.0\\ 
		\hline 
        \multicolumn{4}{l}{\emph{ Setting of SAFE block}}\\
        
        {\textbf{c1:} $K=1$ iteration} & {30.73} & {30.61} &
		\textbf{2.5}\\ 
        {\textbf{c2:} $K=2$ iterations} & {31.04} & {30.77} &
		3.0\\   
		\textbf{c3:} $K=4$ iterations & {31.21} & {30.90} &
		4.0\\ 
  \rowcolor[rgb]{ .9,  .9,  .9}
        \textbf{c4:} $K=6$ iterations & \textbf{31.28} & \textbf{30.97} &
		5.0\\ 
        {\textbf{c5:} + cross-block weight sharing} & 31.07 & 30.81 &
		\textbf{2.5}\\ 
        {\textbf{c6:} non-adaptive flow estimation} & {30.97} & {30.79} &
		5.0\\ 
        {\textbf{c7:} w/o cross-scale weight sharing} & 31.25 & 30.91 &
		6.8\\ 
        {\textbf{c8:} w/o scale selector} & 30.89 & 30.71 &
		5.0\\ 
		\hline 
		\multicolumn{4}{l}{\emph{ Initialization of Feature Extractor}}\\
		\rowcolor[rgb]{ .9,  .9,  .9}
  \textbf{d1:} from scratch & 31.28 & 30.97 &
		5.0\\ 
		{\textbf{d2:} supervised pre-trained~\cite{he2016deep}} & 31.34 & 31.02 &
		5.0\\ 
		{\textbf{d3:} DINO pre-trained~\cite{caron2021emerging}} & \textbf{31.38} & \textbf{31.12} &
		5.0\\ 
		\hline 
		\multicolumn{4}{l}{\emph{Peak Learning Rate $\beta$, number of channels $nc$}}\\
\rowcolor[rgb]{ .9,  .9,  .9}
  \textbf{e1:} $\beta=2\times 10^{-4}$, $nc=80$ & \textbf{31.28} & \textbf{30.97} &
		5.0\\ 
        {\textbf{e2:} $nc=60$} & 31.17 & 30.85 &
		\textbf{2.8}\\
		{\textbf{e3:} $\beta=1\times 10^{-3}$} & 31.22 & 30.91 &
		5.0\\ 
		{\textbf{e4:} $\beta=4\times 10^{-5}$} & 30.94 & 30.62 &
		5.0\\  
		\hline 
		\normalsize
	\end{tabular}
}
	\label{tab:ablation}
\end{table}

We perform extensive ablation studies shown in Table~\ref{tab:ablation}.

\noindent
\textbf{(a) Feature extractor}. The feature extractor in SAFA is built on out-of-the-box ResNet-18~\cite{he2016deep}. We denote the default setting as \{R18: stem, s1, s2\}, indicating that SAFA fuses the features output by the stem and the first two stages. Using only deep features~(\textbf{a1}, \textbf{a2}) brings severe performance degradation. The reason is that more down-sampling layers lead to location confusion~\cite{liu2018intriguing} and make feature alignment difficult. Fusion of multi-level features improves the performance of SAFA~(\textbf{a1} \textsl{vs.} \textbf{a3}). However, advanced models~\cite{xie2017aggregated} (\textbf{a4}, \textbf{a5}) that perform better in high-level tasks may not be suitable for the STVSR task. We found the bottleneck block of ResNet-50 can lead to a performance drop in STVSR. More encoding layers may require additional techniques to preserve spatial information.

\noindent
\textbf{(b) Information aggregation}. SAFA~(\textbf{b1}) computes optical flow for explicit propagation of both image information and features, while implicit propagation~\cite{xiang2020zooming,chan2021basicvsr} usually works on features. The variants~(\textbf{b2, b3}) using only image or feature propagation suffer from clear performance drop.

\noindent
\textbf{(c) Setting of flow estimation}. In the modular design of SAFA, we can perform speed-accuracy trade-off by adjusting the number of flow estimation iterations~(\textbf{c1}-\textbf{c4}). Similar to the finding in the supervised optical flow estimation~\cite{teed2020raft}, the iterative estimation indeed improves the STVSR performance of SAFA. However, sharing weights between different SAFE blocks~(\textbf{c5}) brings about 0.2dB performance drop. Multi-branch selection contributes 0.31/0.18dB improvements and increases the speed~(\textbf{c4} \textsl{vs}. \textbf{c6}, Figure~\ref{fig:time}). The weights between different scales can also be shared, which reduces model parameters and makes SAFA more effective~(\textbf{c4} \textsl{vs}. \textbf{c7}). Directly optimizing the scale selector~(\textbf{c8}) rather than associating it with the inputs~\cite{liu2018darts} brings a performance drop, which confirms the importance of input-adaptive scale selector.

We study how SAFE block is benefited from selecting model branches. We perform $8\times$ time scale and $4\times$ space scale STVSR on the X4K1000FPS dataset~\cite{sim2021xvfi}. We bilinearly down-sample the original frames to get video clips of different resolution. As shown in Table~\ref{tab:xvfi}, SAFA tends to select larger scales when processing videos in higher resolution. When fixing scale=$1$, the receptive field may not cover large motions, bringing performance degradation.

\begin{table}[]
\caption{\textbf{Statistics of scale selection} for different input resolution and performance impact.}
\resizebox{0.47\textwidth}{!}{
\begin{tabular}{l|cccc}
\hline
Input Resolution      & 4K   & 2K   & 540p & 270p \\ \hline \hline
Ratio of scale=1/4      & 0.71 & 0.58 & 0.49 & 0.41 \\ 
Ratio of scale=1/2      & 0.17 & 0.18 & 0.15 & 0.16 \\ 
Ratio of scale=1      & 0.12 & 0.24 & 0.36 & 0.43 \\ \hline
Fix scale=1 PSNR & 27.05    & 25.50     & 23.55     & 21.62     \\ 
SAFA PSNR        & \textbf{32.28} & \textbf{29.02}    & \textbf{24.90}     & \textbf{22.11}      \\ 
VideoINR~\cite{chen2022videoinr} PSNR         & 29.70 & 26.64    & 23.09     & 20.56     \\ \hline
\end{tabular}
\label{tab:xvfi}
}
\end{table}

\noindent
\textbf{(d) Benefited from pre-trained methods.} Based on out-of-the-box CNN architectures, the feature extractor of SAFA can benefit from the research on other tasks. Due to the huge difference between low-level and high-level vision tasks, using the paradigm of fine-tuning pre-trained models in low-level tasks is still challenging~\cite{kotar2021contrasting}. SAFA can utilize pre-trained models to improve its STVSR performance, especially using unsupervised pre-trained parameter weights~(\eg, DINO~\cite{caron2021emerging})~(\textbf{d3}). For a fair comparison with the comparison methods, we initialize from scratch the parameters of the feature extractor in SAFA. We reported these results to show potential future improvements for inspiring future work. 

\noindent
\textbf{(e) Choice of hyper-parameters}. 
We set the number of channels $nc=80$ in SAFA. Reducing model size could increase speed, but also decrease performance~(\textbf{e2}). Choosing a larger or smaller peak learning rate degrades the performance~(\textbf{e3},\textbf{e4}).

\section{Conclusion}
We developed a powerful network for STVSR. Specifically, we proposed a Scale-Adaptive Flow Estimation (SAFE) block to perform scale selection for accurate motion modeling. We further introduced an iterative estimation scheme into SAFA to perform scale-adaptive feature propagation and fusion for efficient STVSR performance. Experimental results demonstrated that SAFA outperforms predecessors comprehensively. Extensive ablation studies confirm the efficacy of our approach. We envision SAFA as a solid foundation for future research. Further direction includes exploring the potential of SAFA on the real-world STVSR task and more advanced feature extractor. 

\noindent\textbf{Limitation.} Limited by the studied topic, our work may not cover some  scenarios. Firstly, following the setting of VideoINR~\cite{chen2022videoinr}, SAFA focuses on using two input images. For multi-frame inputs, SAFA could benefit from more implicit feature propagation~\cite{xiang2020zooming,kalluri2020flavr,chan2021basicvsr}. SAFA is suitable for streaming systems as it does not depend on distant future frames. Secondly, our experiments are done with PSNR and SSIM as quantitative metrics to objectively measure the capability of the models~\cite{blau2018perception}. SAFA can be readily changed to use the perceptual losses to cater the perception. Thirdly, SAFA may not tackle the diverse degradation that exists in real-world videos currently. To this end, it is necessary to train SAFA using additional training data and specialized training pipelines designed for real-world STVSR~\cite{zhang2020ntire}.

\noindent
\textbf{Acknowledgements}. Jun Xu is partially supported by the National Natural Science Foundation of China (No. 62002176 and  62176068), and the Open Research Fund (No. B10120210117-OF03) from the Guangdong Provincial Key Laboratory of Big Data Computing, The Chinese University of Hong Kong, Shenzhen.

{\small
\bibliographystyle{ieee_fullname}
\bibliography{egbib}
}
\onecolumn
\newpage

\begin{figure*}[h]
	\centering
	\includegraphics[width=15cm]{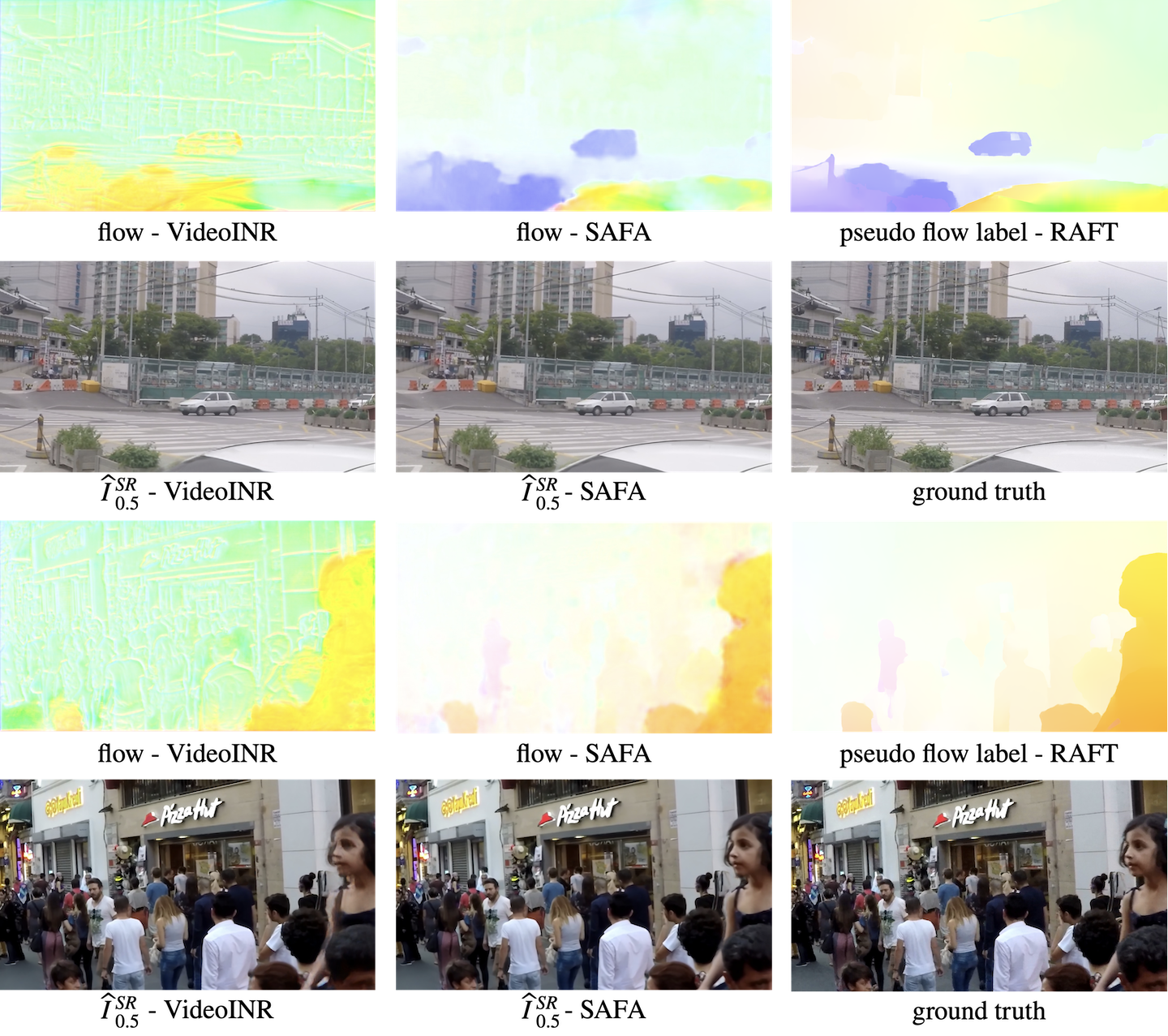}
	\caption{\textbf{Visualization of the intermediate flow estimated by SAFA. The pseudo label is obtained using RAFT~\cite{teed2020raft}}.  }\label{fig:flow}
\end{figure*}

\paragraph{Societal Impact.} STVSR methods can remove and synthesize frames, and the processed video may reflect different facts. This can be used by artists as a creative tool, but it may be used inappropriately. The related editing detection and reliability verification methods require further research.

\section{Video Effect and Failure Case}
The results of this Appendix are generated on the GoPro dataset~\cite{nah2017gopro}. The video demo is attached. We mainly compare SAFA with VideoINR~\cite{chen2022videoinr} because it has state-of-the-art quantitative results. By observing the video results, we find that SAFA has an advantage over VideoINR~\cite{chen2022videoinr} mainly when the object or camera motion is large. In addition, the recovery of regions with complex textures using SAFA is also basically better. SAFA still has two types of artifacts that affect the perception. 1) At the border of the video, some objects will move out of the screen, similar to VideoINR. At this time, it is difficult for the model to learn a reasonable transition, showing the fading in and fading out effects. Designing inpainting components may be able to remedy this shortcoming. 2) In some repetitive texture areas, such as fences, floor tiles, etc., the model may distort the lines. Such artifacts may be counteracted by adding smoothness constraints to the flow fields.

\begin{figure*}[h]
	\centering
	\includegraphics[width=10cm]{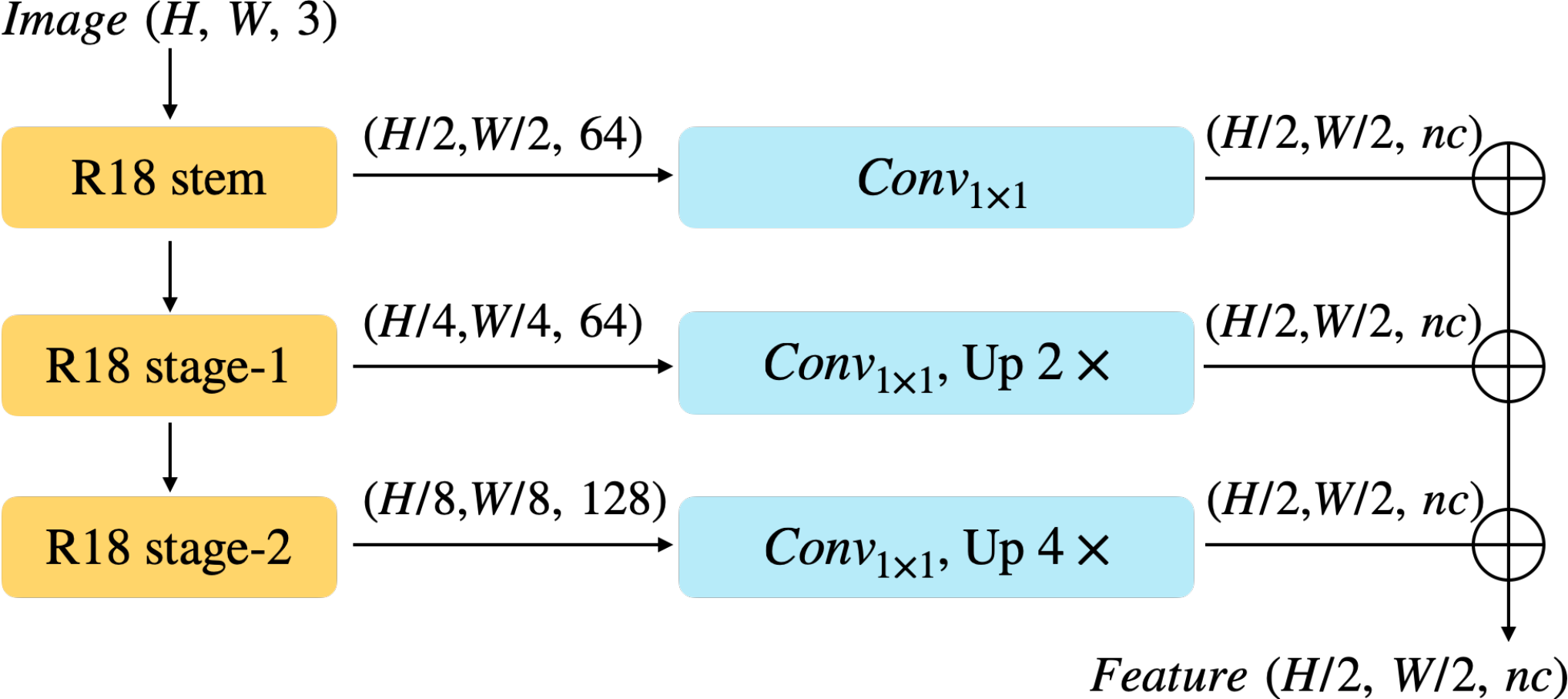}
	\caption{\textbf{Architecture of R18 Feature Extractor.}
 We use a $1\times 1$ convolutional layer and bilinear up-sampling to adjust the number of channels and feature map size.}\label{fig:r18}
\end{figure*}

\begin{figure*}[th]
	\centering
	\includegraphics[width=15cm]{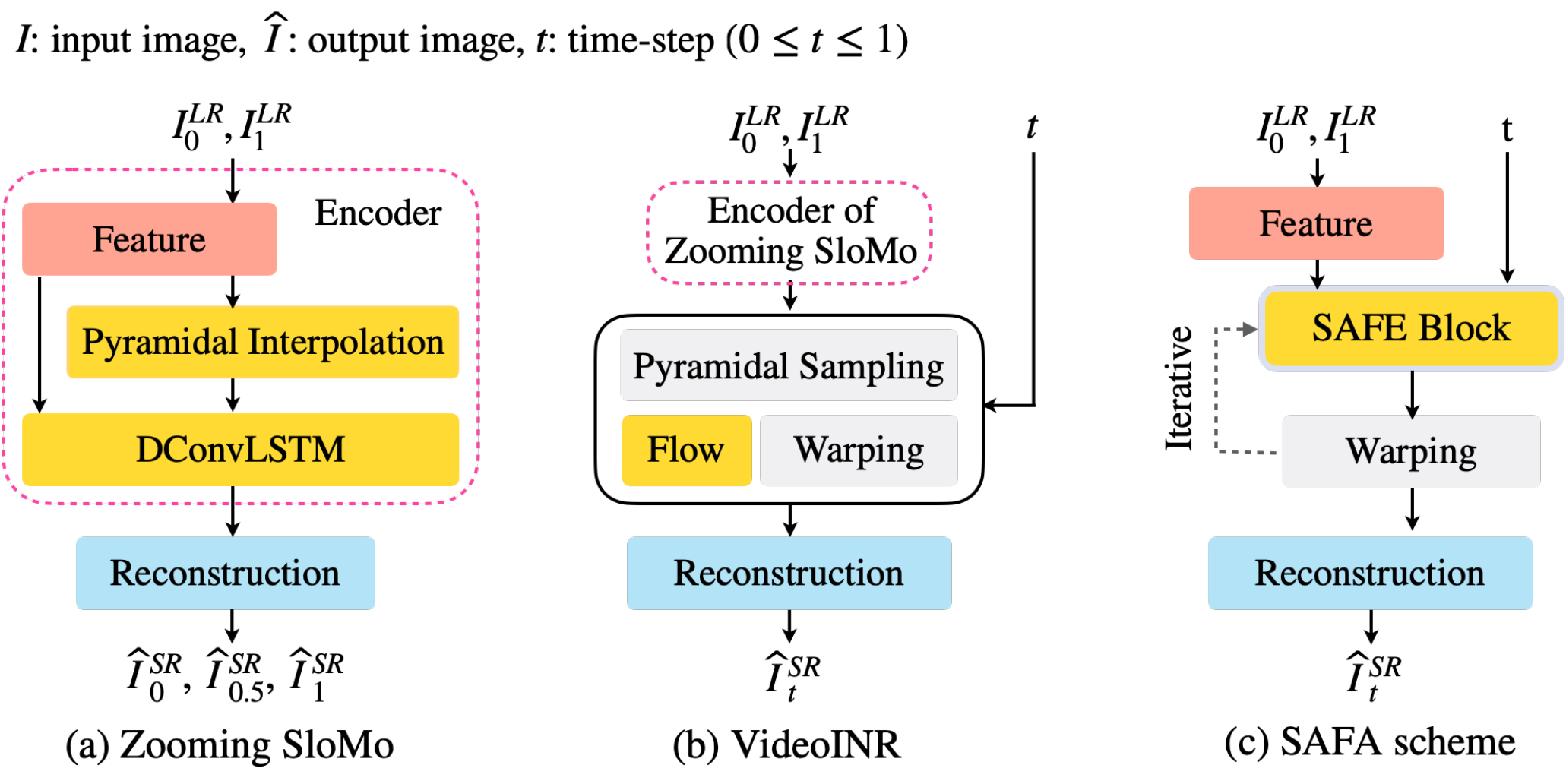}
	\caption{\textbf{Comparison of different structures.} We color blocks with similar functions the same. }\label{fig:arch}
\end{figure*}

\section{Analysis of Intermediate Flow}

Our proposed SAFA explicitly uses intermediate flows for feature propagation. To confirm that the architecture of SAFA indeed learns an optical flow-like representation, we show the visualization of approximated intermediate flow in Figure~\ref{fig:flow}. We use the state-of-the-art optical flow model, pre-trained RAFT-things~\cite{teed2020raft}, to generate pseudo flow labels on the ground truth image and observe the difference. 

\begin{figure*}[t]
	\centering
	\includegraphics[width=14cm]{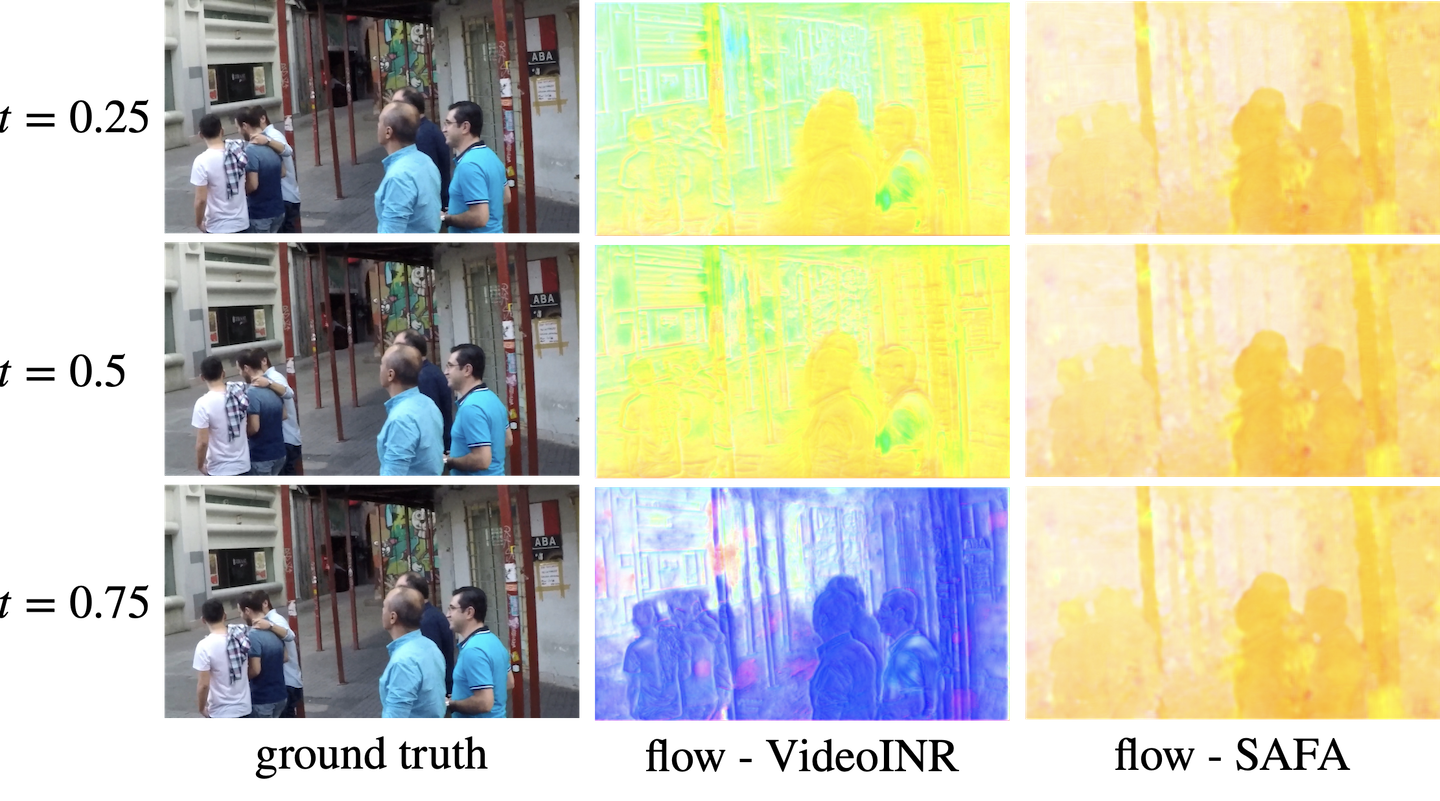}
	\caption{\textbf{Visualization of the estimated flow in different time-step.} }\label{fig:flow3}
\end{figure*}

We show that the intermediate flow estimated by SAFA is similar to the pseudo flow label of RAFT~\cite{teed2020raft}. From the appearance, the flow pseudo label has sharper boundaries and is cleaner. This is mainly due to the difference in the definition of task-oriented flow~\cite{xue2019video} and optical flow. On the other hand, it is also partly due to the low resolution of the input of STVSR. Whereas the VideoINR~\cite{chen2022videoinr} estimated flow is quite different. We can only see similar object edges. This demonstrates that the Zooming Slomo~\cite{xiang2020zooming} encoder in VideoINR~\cite{chen2022videoinr} has already undertaken part of the feature alignment. We depict the architecture of these previous methods and SAFA in Figure~\ref{fig:arch}. The encoder is entangled with flow estimation. We argue that an explicit modular structure is important for designing efficient models. The intrinsic relationship of estimated flows at different time-steps is shown in Figure~\ref{fig:flow3}. It can be seen that SAFA maintains good consistency. 

\section{Analysis of Fusion Map and Refinement}

\begin{figure*}[t]
	\centering
	\includegraphics[width=14cm]{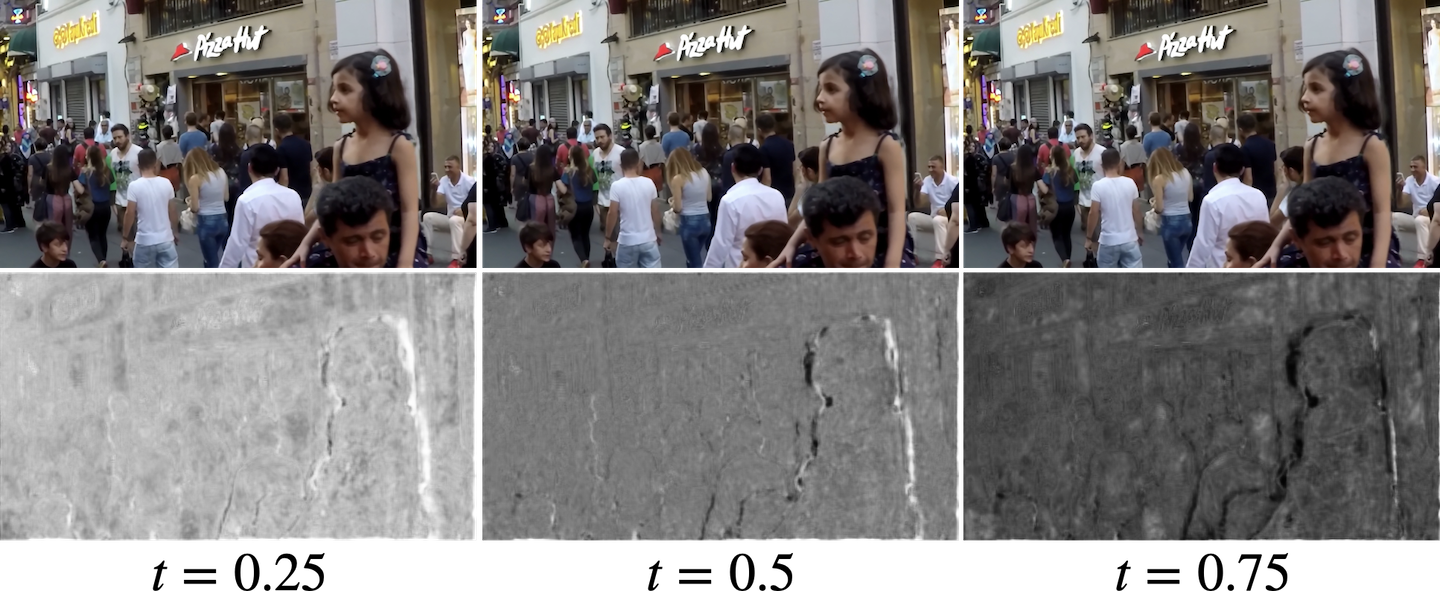}
	\caption{\textbf{Visualization of generated frames and the corresponding occlusion map.} }\label{fig:mask}
\end{figure*}

\begin{figure*}[h]
	\centering
	\includegraphics[width=12cm]{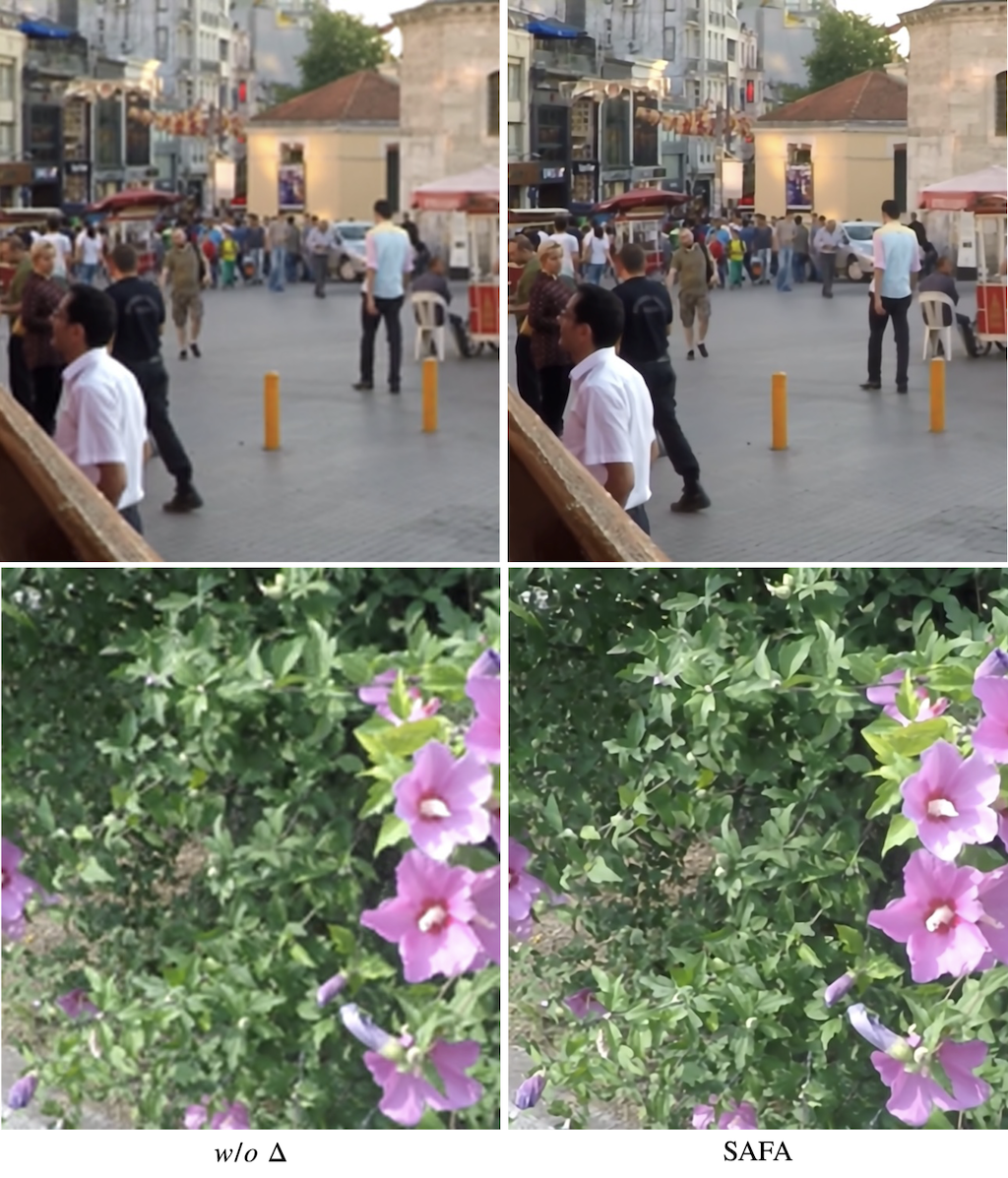}
	\caption{\textbf{Visualization of generated frames with/without $\Delta$. They have a noticeable difference in image sharpness.} }\label{fig:delta}
\end{figure*}

In SAFA, the formula that produces the final result is:
\begin{equation}
\widehat{I}_t^{SR} = [\textbf{m}\odot\widehat{I}_{t\leftarrow0} + (1 - \textbf{m})\odot\widehat{I}_{t\leftarrow1}] + \Delta,
\end{equation}
where $m$ is usually called ``fusion map" or ``occlusion map"~\cite{jiang2018super,niklaus2020softmax,huang2022rife}. For non-occluded regions, it is used to weigh between the two results. Intuitively, when the time-step $t$ is smaller, $m$ is closer to $1$ (visualized as white), making the model consider more the results from $I_0$. The occluded area often appears at the edge of the moving objects, and the model will choose one of the two results adaptively. The visualization is shown in Figure~\ref{fig:mask}. Because $I_0$ and $I_1$ are both low-resolution images, it is intuitively impossible to obtain high-resolution images simply by warping and fusing them. The visual effect without feature-based refinement ~$\Delta$~(reconstruction model) is shown in Figure~\ref{fig:delta}.

\section{Comparison with Pyramidal Design}
Scale-selection increases the flexibility (and thus reduces the burden) of hand-crafted pyramid design models. On the other, we can share parameters at different scales~(Table 3, \textbf{c7}). We fix the scale of the 6 blocks to (0.25, 0.25, 0.5, 0.5, 1, 1) to construct a pyramidal-like structure. It cannot scale adaptively during inference.

\begin{table*}[h]
\centering
\resizebox{0.5\textwidth}{!}{\begin{tabular}{lccc}
		\hline
		\multicolumn{1}{c}{\multirow{2}{*}{Supplementary Table 1}} & \multicolumn{1}{c}{GoPro} & \multicolumn{1}{c}{Adobe240} & \multicolumn{1}{c}{\# Param} \\ 
		
		\multicolumn{1}{c}{}                         & \multicolumn{1}{c}{PSNR}   & \multicolumn{1}{c}{PSNR}     & \multicolumn{1}{c}{(M)}            \\ \hline \hline
		\textbf{f1:} SAFA & \textbf{31.28} & \textbf{30.97} &
		5.0\\ 
		\textbf{f2:} Manually Set Scale & {31.04} & {30.73} &
		{5.0}\\ \hline
		\normalsize
	\end{tabular}
}
\vspace{-2em}
\end{table*}

\section{Specific Training Cost}
For these methods for comparison, we u se open-sourced codes. On four Pascal TITAN X GPUs, TMNet~\cite{xu2021tmnet} and VideoINR~\cite{chen2022videoinr} take about 200 hours and 140 hours to train, respectively. While SAFA takes only 50 hours. The three methods use the same number of training iterations, TMNet~\cite{xu2021tmnet} outputs 7 frames per iteration, while VideoINR~\cite{chen2022videoinr} and SAFA output 3 frames at each forward pass. This is one reason why the training overhead of TMNet~\cite{xu2021tmnet} is higher. In other words, TMNet undergoes more data iterations.

\end{document}